\documentclass[letterpaper, 10pt, conference]{IEEEtran}
\IEEEoverridecommandlockouts

\usepackage{cite}
\usepackage{amsmath,amssymb,amsfonts}
\usepackage{algorithmic}
\usepackage{graphicx}
\usepackage{textcomp}
\usepackage{booktabs}
\usepackage{xcolor}
\usepackage{url}
\def\BibTeX{{\rm B\kern-.05em{\sc i\kern-.025em b}\kern-.08em
    T\kern-.1667em\lower.7ex\hbox{E}\kern-.125emX}}

\usepackage{hyperref}
\hypersetup{colorlinks}

\newcommand{\reward}[1]{R_{\text{#1}}}

\begin{document}
\title{\LARGE \bf
    Deep Reinforcement Learning for Continuous Docking Control of Autonomous Underwater Vehicles: A Benchmarking Study
}

\author{
    Mihir Patil$^{1,2}$
    \and
    Bilal Wehbe$^{1}$
    \and
    Matias Valdenegro-Toro$^{1}$
    \thanks{
        $^{1}$Bonn-Rhein-Sieg University of Applied Sciences, 53757 Sankt Augustin, Germany
        {\tt\small mihir.patil@smail.inf.h-brs.de}%
        
        $^{2}$German Research Center for Artificial Intelligence, 28359 Bremen, Germany
        {\tt\small matias.valdenegro@dfki.de, bilal.wehbe@dfki.de}}%
}

\maketitle

\begin{abstract}
    Docking control of an autonomous underwater vehicle (AUV) is a task that is integral to achieving persistent long term autonomy. This work explores the application of state-of-the-art model-free deep reinforcement learning (DRL) approaches to the task of AUV docking in the continuous domain. We provide a detailed formulation of the reward function, utilized to successfully dock the AUV onto a fixed docking platform. A major contribution that distinguishes our work from the previous approaches is the usage of a physics simulator to define and simulate the underwater environment as well as the DeepLeng AUV.
    We propose a new reward function formulation for the docking task, incorporating several components, that outperforms previous reward formulations.
    We evaluate proximal policy optimization (PPO), twin delayed deep deterministic policy gradients (TD3) and soft actor-critic (SAC) in combination with our reward function. Our evaluation yielded results that conclusively show the TD3 agent to be most efficient and consistent in terms of docking the AUV, over multiple evaluation runs it achieved a 100\% success rate and episode return of $10667.1 \pm 688.8 $. We also show how our reward function formulation improves over the state of the art.
\end{abstract}


\section{Introduction}
Autonomous underwater vehicle (AUV) docking has become an integral part in marine applications that require prolonged deployment periods, such as exploring deep oceans, seafloor environment monitoring, underwater archaeology. In such oceanic missions, it helps to achieve long-term autonomy by facilitating functions such as downloading data, uploading a new mission plan and recharging the battery, while also reducing expensive recovery costs. However, docking control is yet a challenging task, the highly non-linear vehicle dynamics, contacts between the vehicle and the docking station as well as the presence of environmental disturbances pose considerable issues to any potential docking maneuver.

A vast majority of studies on docking control, combine acoustic, electromagnetic fields or vision based strategies with classical control algorithms, namely proportional integral derivative (PID), sliding mode (SM) or pure pursuit control. Yet the distinct non-linear nature of underwater vehicle hydrodynamics with its uncertainties that are challenging to parameterize, in conjunction with the environmental disturbances that are quintessential of an aquatic environment, restricts in the utility of such classical control approaches.

This has led the research to explore of artificial intelligence (AI) as a viable alternative, with deep reinforcement learning (DRL) emerging as a major driving force towards realizing persistent autonomy. DRL has shown promising results for robotic applications such as indoor navigation and object manipulation. It's ability to learn control policies from raw and high-dimensional sensory inputs is an asset that could potentially benefit AUV docking control applications in achieving adaptive behaviour for real-time problem solving. In this work, we propose to explore and evaluate the possibility of using state-of-the-art model-free DRL algorithms for docking control of an AUV onto a fixed underwater platform.

This work explores AUV docking in the continuous domain by using a state-space defined by the pose, velocity and thruster values of the AUV. The goal is to learn a control policy which provides real valued thruster outputs required to control the AUV. One of the main issues tackled here is the realization of a reward function that can be used to successfully dock the AUV onto a fixed docking platform. A major contribution that distinguishes our work from the previous approaches that use DRL, is the usage of a physics simulator to define and simulate the underwater environment as well as a model of an AUV ``DeepLeng'' \cite{hildebrandt2020epi} that is being currently developed at the labs of DFKI-Robotics Innovation Center in Bremen, Germany.

\begin{figure}[t]
    \centering
    \includegraphics[trim=0 80 0 80,clip,scale=0.3]{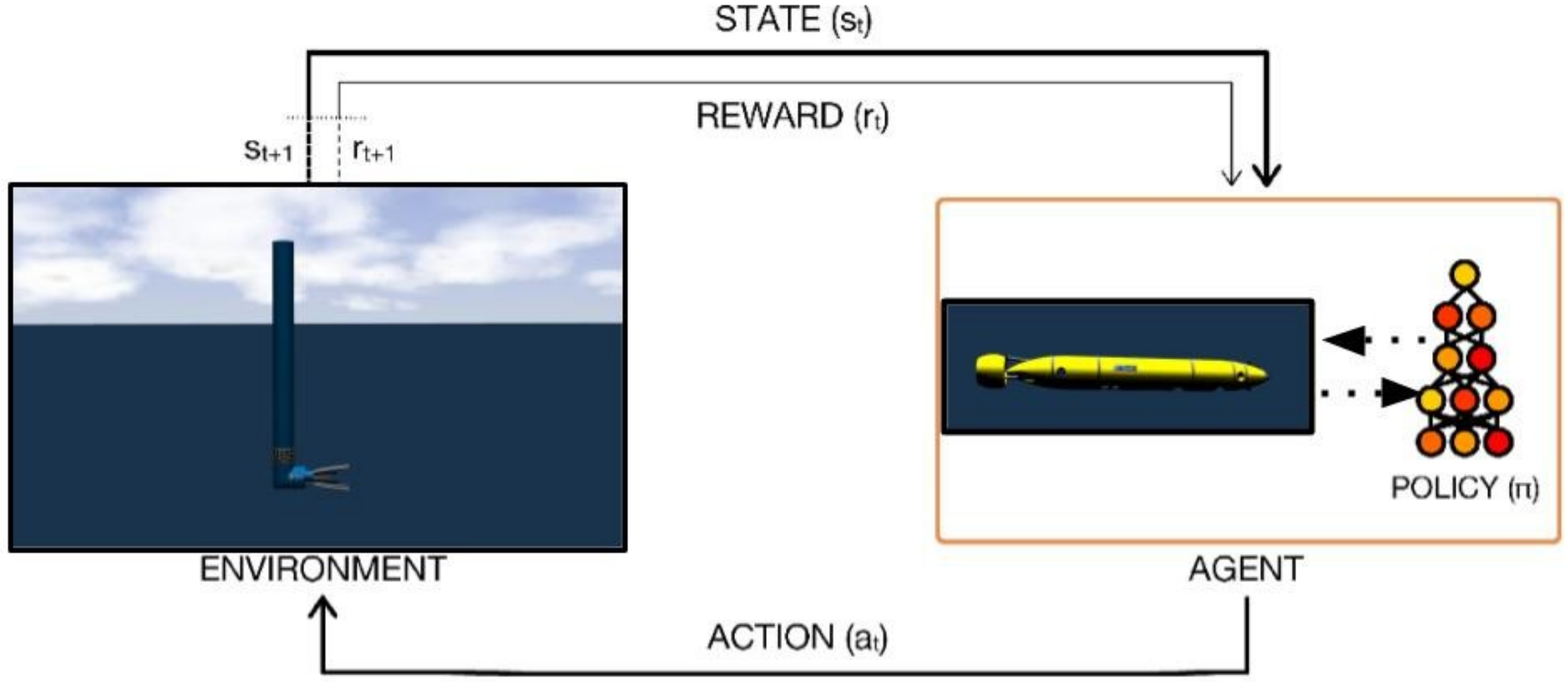}
    \caption{The perception-action-learning loop used to realize the docking control. At each time step the agent at time perceives state $s_t$ from the environment. It executes an action $a_t$ using its policy thereby proceeding the next state $s_{t+1}$ and receives a reward $r_{t+1}$. Finally the tuple ($s_t$, $a_t$,$s_{t+1}$, $r_{t+1}$) to learn an optimal policy that maximizes the cumulative expected reward.}
    \label{fig:basic rl}
\end{figure}

\section{State of the art}
Approaches such as PID, optimal control, pure pursuit control and model based control have been the mainstay in underwater robotics. The works published in \cite{teo2012robust, GUO20032137, SMITH1993318} employ fuzzy logic together with PID or SM to approximate the system's control parameters with varying success, yet the intricate knowledge required to craft the fuzzy rules and membership functions has hindered it's widespread adoption \cite{carlucho2018adaptive}. These approaches have been slowly evolving over time to obtain adaptive controllers that can sufficiently handle the high variability and hostility of the aquatic environment. The authors in \cite{lee2003visual, maire2009vision} incorporate Aruco markers or led arrays \cite{park2009experiments} in combination with PID or pure pursuit controllers to achieve visual servoing based AUV docking. Yet the low light, murky state of the underwater environment mitigates the utility of such an approach. Particularly, in the case of pure pursuit controllers \cite{mcewen2008docking} which do not take into account ocean currents and are hence prone to be easily blown off course.

Model predictive control is another promising avenue for developing adaptive docking controllers \cite{oh2002homing, park2009experiments} but inaccurate models as well as model uncertainties have made this approach less desireable \cite{wu2018depth}. The difficulty in parameterizing the hydrodynamics and irregularities of the underwater environment \cite{carlucho2018adaptive} has restricted the use of classical control approaches outside idealized conditions.

Notable examples of the advancements in Deep RL for AUV control are  \cite{Anderlini_2019, carlucho2018adaptive, carlucho2018auv, yu2017deep, wu2018depth}. The authors of \cite{carlucho2018auv} used DDPG to perform 5 degrees of freedom (DOF) waypoint navigation and derive an end-to-end position control policy using the AUV's relative position and velocity as input while producing the thruster signals necessary to reach a dynamic goal. The approach was tested in simulation, however, the authors do not provide any details regarding the simulator used.

The work of \cite{wu2018depth} evaluates performance of model-free deterministic policy gradient (DPG) on tasks such as constant depth control in the xz-plane, curved depth tracking and seafloor tracking according to different target trajectories and information, while 
assuming constant surge velocity.

The research in \cite{yu2017deep} used DDPG to perform tracking control for both straight and curved trajectories of an AUV in the horizontal plane with a state-space defined by the position and velocity, whereas the actions were represented by motor torques.

The work presented in \cite{Anderlini_2019} is the only one focusing on underwater docking of an AUV. This work evaluates the DDPG and DQN algorithms against a PID and an optimal control approach. This study concludes that DRL approaches are 5 times faster than traditional optimal control approaches at deployment time, while achieving performance similar to optimal control.

The issues illustrated in the studies are reflective as a whole on the predicament of AUV control using DRL. In this work we aim to fill in the gap of a reliable benchmarking study in that is currently lacking in the field of DRL based AUV docking. For our study we make use of the DeepLeng AUV (under development at DFKI-RIC, Bremen) and  attempt to learn the full range of motions available to it, in an environment that is modelled with a physics simulator.

\section{Methodology}
\subsection{Software pipeline}
\begin{figure}[t]
    \centering
    \includegraphics[scale=0.45]{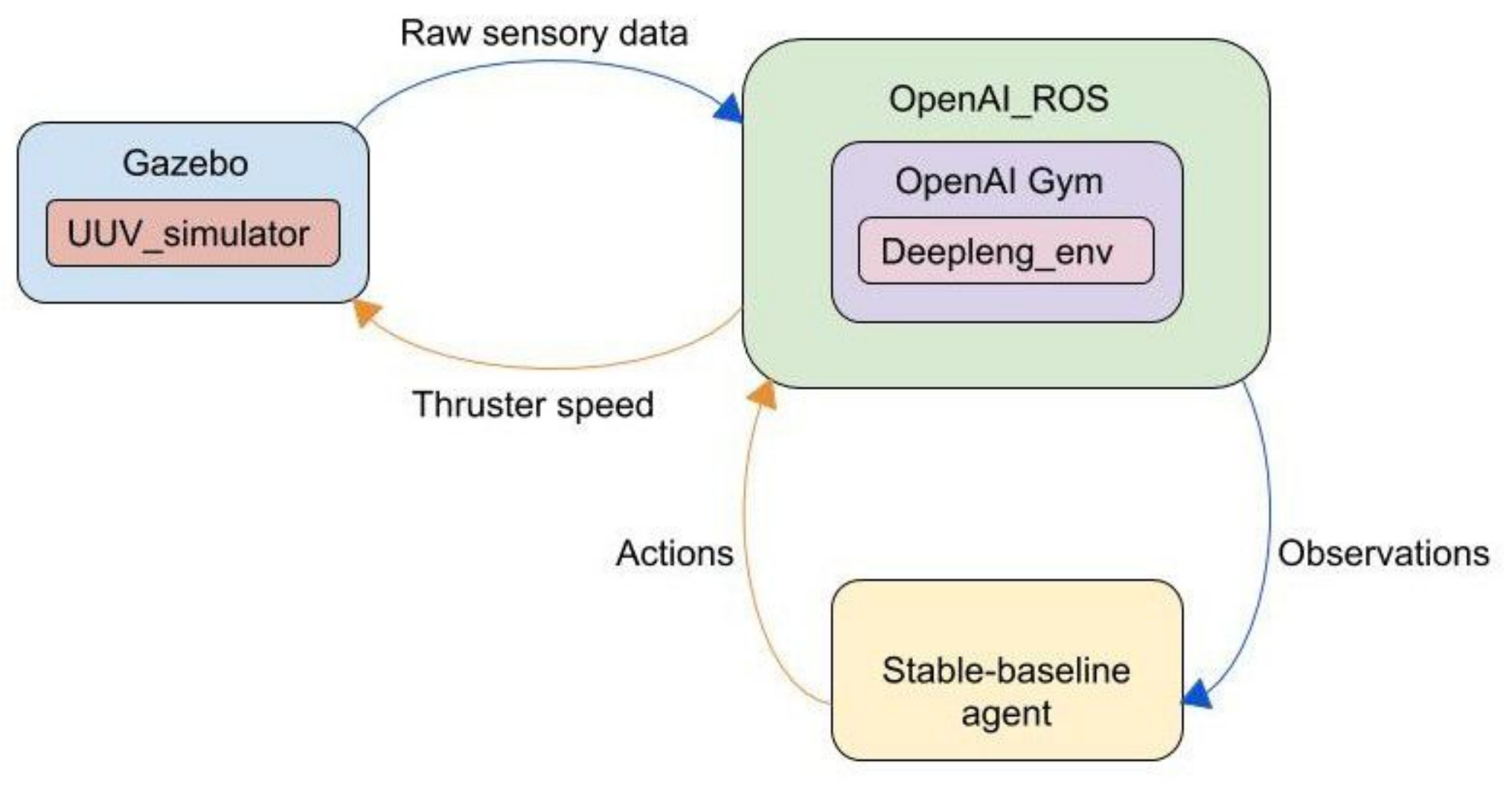}
    \caption{Simplified overview of the software pipeline depicting the end-to-end learning architecture followed in this work.}
    \label{fig:approach}
\end{figure}

The core concept of this work is to leverage DRL and obtain adaptive, intelligent agents that can perform the task of AUV docking control in a 2D space. To achieve this goal we make use of the ROS ecosystem in combination with Gazebo and UUV\_Simulator\footnote{\url{https://github.com/uuvsimulator/uuv_simulator}}. Moreover, taking advantage of ROS's modularity we propose on implementing each component of our pipeline as stand-alone packages which can be modified individually without affecting the harmony of the entire pipeline and the associated training process. An graphical overview of the proposed pipeline is shown in figure \ref{fig:approach}. 

To harness the full capability of DRL we conform to the end-to-end problem formulation for our use case. Wherein, raw real valued sensor data from the simulator setup, is provided as observation to the learning agents which are deployed using the stable-baselines framework\footnote{\url{https://github.com/hill-a/stable-baselines}}. These agents use the multi-layer perceptron (MLP) model of a neural network (NN) as function approximators to obtain control signals in terms of individual speeds (actions of the DRL agent) for all the thrusters present on the AUV.

\subsection{Simulation Scenario}
\begin{figure}[t]
    \centering
    \includegraphics[scale=0.25]{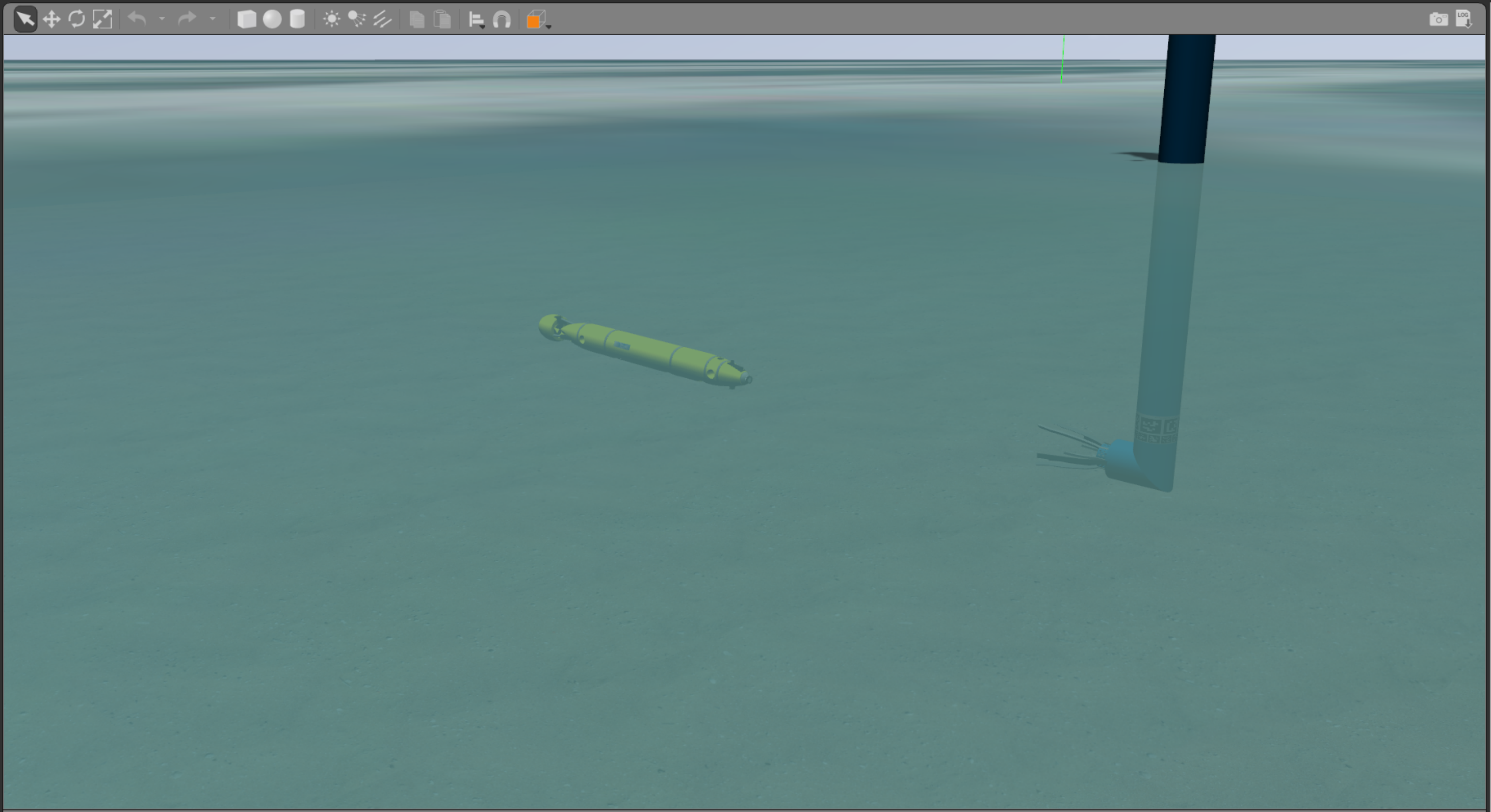}
    \caption{Illustration of our simulation scenario.}
    \label{fig:scenario}
\end{figure}

A simulation environment was created using the open-source package UUV\_Simulator.
Fig.~\ref{fig:scenario}. The AUV is modeled as a free-floating body that is maneuverable in the horizontal plane using 1 thruster for surge and 2 thrusters for sway and yaw. note that roll, and pitch are set to zero, and the depth of the vehicle is set to match the depth of the docking station. The docking station is fixed to the origin of the simulation workspace where the center of the docking cone is completely submerged at 2m depth. We rely on the north east down (NED) coordinate frame commonly used in marine applications for the purpose defining the geographic reference frame.

To interface between an agent and the simulation, and environment using openai-gym was developed that allows the agent to read the robot's state and forward the action commands generated by the agent's policy. We define the following parameters for our simulations:

\begin{itemize}
    \item \textbf{State-space:} We define our state space based on the pose, velocity and thruster speeds of the AUV at any given point in time, such that the observation is a 9 dimensional vector (following the SNAME convention):  $$\big[\begin{smallmatrix} x, & y, &\psi, & u, & v, & r, & n_1, & n_2, & n_3 \end{smallmatrix}\big]$$ which describes the current state of the AUV and it's thruster speeds ($n_1, n_2, n_3$).
    
    \item \textbf{Initial states:} The AUV's initial state is set randomly at the start of every episode. This helps to avoid any bias arising from starting in a single fixed position. To ensure that the initial position of the AUV is not directly within the docking station or too close to it, we inhibit the starting positions along the x and y axes to be within 2 meters of the maximum workspace limit (9 meters).
    
    \item \textbf{Terminal states:} The terminal states are cases where the agent has either driven the AUV outside the stipulated workspace bounds or driven it expertly to the goal position which is located at the docking station. Consequently we make use of these two situations to formulate our finite-horizon episodic control task.
    
    \item \textbf{Action space:} The action space is defined as a 3-dimensional vector $\big[\begin{smallmatrix}n_1, & n_2, & n_3 \end{smallmatrix}\big]$, representing the rotational speed of each thruster.
\end{itemize}

\subsection{Reward formulation}
We formulate our reward function to avoid sparsity by associating every point in the state-space with a reward value based on it's relative pose with respect to the goal (docking station pose). This reward is reinforced with a penalization on the thruster utilization with the intention of efficient energy consumption. To avoid the issue of reward gathering via infinite exploration of the state-space our reward is inherently negative in nature such that negative reward keeps reducing in magnitude as the AUV approaches the docking station and only turns into a large positive value upon reaching the docking station. Such a formulation helps mitigate the possibility of a local minima to a certain extent. Moreover the two components of our reward (pose and thruster utilization) are weighted separately as the significance of each is not always the same and this also allows the agent to implicitly learn the more desireable of two possible states given a set of actions.

\subsubsection{Distance}
The most basic reward function is defined by the relative distance between the goal and current position of the AUV. 
\begin{equation}
\reward{dist} = - w_d \sqrt{(c - g)^2}
\label{eqn:edist}
\end{equation}
where $c$ and $g$ are the current and goal position vectors respectively, represented by $[x, y]$. Whereas $w_d$ represents the weight given to the distance component.

\begin{figure}[t]
    \centering
    \includegraphics[scale=0.45]{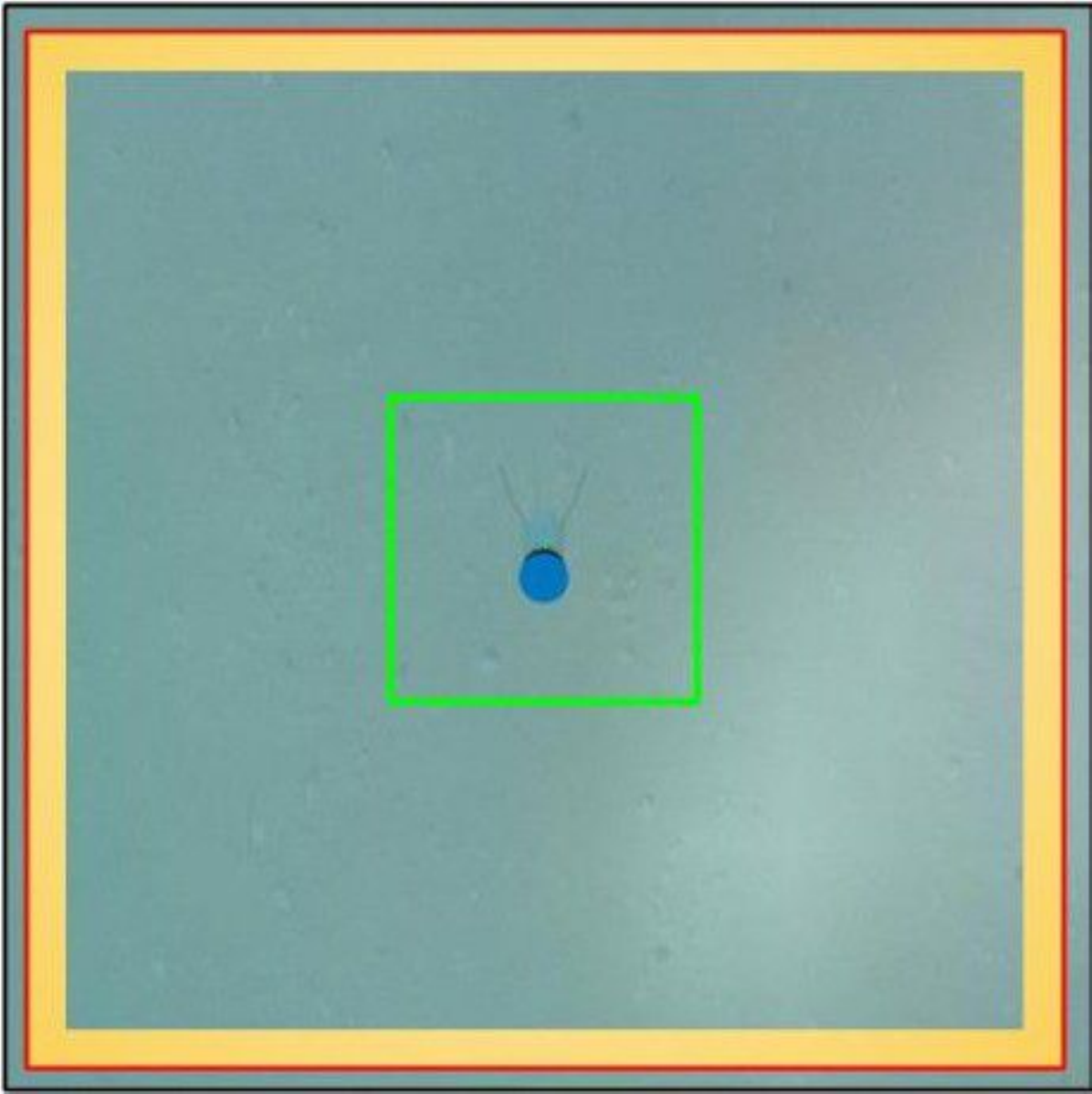}
    \caption{Experimental setup when the reward at each timestep is based solely on the relative distance between the AUV and the goal, which is shown here as a green square surrounding the docking station.}
    \label{fig:dist}
\end{figure}

\subsubsection{Thruster utilization}
The primary factor affecting the velocity and energy consumption of the AUV are the thrusters. The AUV's velocity depends directly on the rotations per minute (rpm) of the thrusters, a higher rpm over a sustained period of time directly translates into higher velocity for the AUV, but also reduces it's operational window. Therefore a trade-off is used wherein the thruster utilization is penalized with smaller values, so that the thrusters may be used but there is a proportional penalty for excessive utilization. Adding thruster penalization to the distance component leads to a more stable learning curve.
\begin{equation}
\reward{thrust} = -\sum_{i}w_{th_i}(\lvert n_i \rvert)
\label{eqn:thruster}
\end{equation}
here $w_{th_i}$ represent the weight associated to the utilization of a given thruster, and $n_i$ the thusters' rpm.

\subsubsection{Orientation}
\begin{figure}[t]
    \centering
    \includegraphics[scale=0.33]{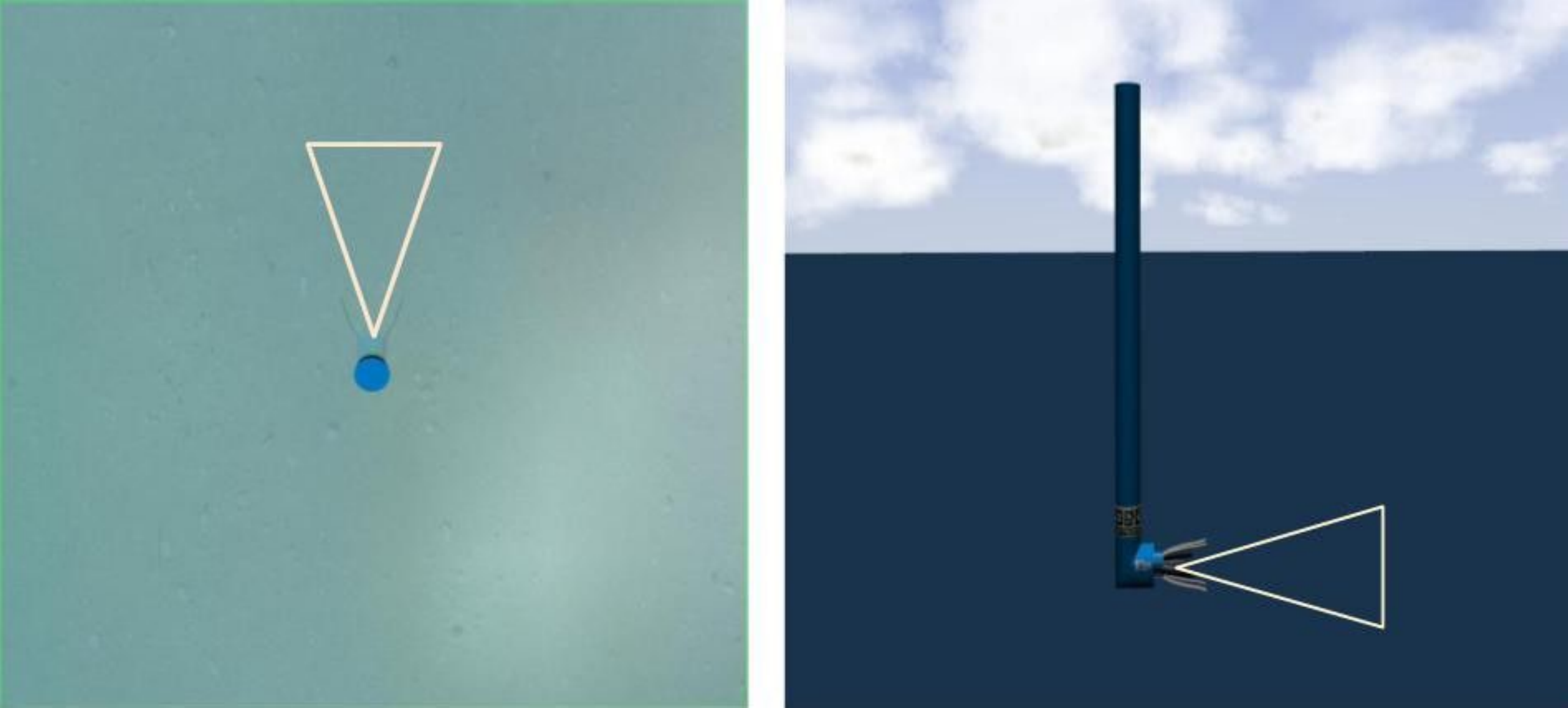}
    \caption{A top and side view of the geometric region for the orientation reward, this region resembles a pyramid in 3D space and a triangle in 2D space.}
    \label{fig:cone}
\end{figure}

Achieving the correct goal orientation can be a challenge when the initial pose of the AUV is behind the docking station. As the negatively weighted distance component seen in equation \ref{eqn:edist} drives the AUV towards the goal, there is a possibility of collision between the vehicle and the docking station. To avoid this scenario we define a geometric component for our reward function, unambiguously this is a region in front of the docking station opening (shown Fig.~\ref{fig:cone}). Within this region we reduce the weight associated with the distance component, thereby creating a region of significantly lower negative reward and motivating the learning agent to drive the AUV through this region rather than through the surrounding area where the negative reward remains quite high. We also utilize this region to calculate the orientation reward as seen in equation \ref{eqn:orient}. The reason for limiting the orientation reward to this triangular region, is to prevent the AUV from prematurely aligning with the desired orientation and therefore drive backwards towards the goal position. For further discussions in this chapter we refer to this geometric region as the docking pyramid.

\begin{equation} \hspace*{-0.25cm}
R_{align}  = \left\{ \begin{aligned}
& 0, \hspace*{2.6cm} \text{outside docking pyramid}\\
& - w_\psi(\lvert \psi_c - \psi_g \rvert) - w_{y}(\lvert y_c - y_g \rvert), \\
&\hspace*{3cm}\text{inside docking pyramid}\ 
\end{aligned} \right. \, \\
\label{eqn:orient}
\end{equation}

Where $w_\psi, w_y$  represent the weights given to the absolute difference between the AUV's current ($c$) and goal ($g$) values for yaw, and the position along the y axis respectively. As described earlier our reward function includes the orientation component only when the agent has managed to drive the AUV into the docking pyramid (3D case) or docking triangle in the 2D case. If the AUV is not in the docking pyramid the orientation reward is set to 0 and our reward function will then constitute of only the distance and thruster penalization components shown in equation \ref{eqn:edist} and equation \ref{eqn:thruster} respectively.

Thus our continuous reward function (equation \ref{eqn:rewfunc}) is composed of three components
\begin{equation}
\reward{continuous} = \reward{dist} + \reward{thrust} + \reward{align} 
\label{eqn:rewfunc}
\end{equation}

Through these principal ideas the final reward function is then defined by just a simple summation of the continuous reward shown in equation \ref{eqn:rewfunc} and the fixed rewards utilized for the terminal states as shown in equation \ref{eqn:finrew}.

\begin{equation}
\reward{final} =
\begin{cases}
\reward{goal} + \reward{continuous}, & \text{goal reached}\ \\
\reward{violation}, & \text{workspace violation}
\end{cases}
\label{eqn:finrew}
\end{equation}

We use values $\reward{goal} = 10000$ and $\reward{violation} = -25000$ during tuning of reward weights, and $\reward{goal} = 15000$ and $\reward{violation} = -25000$ for the final agent evaluation.

\section{Experimental Evaluation}
\label{sec:setup}
\begin{figure}[ht]
    \centering
    \includegraphics[scale=0.48]{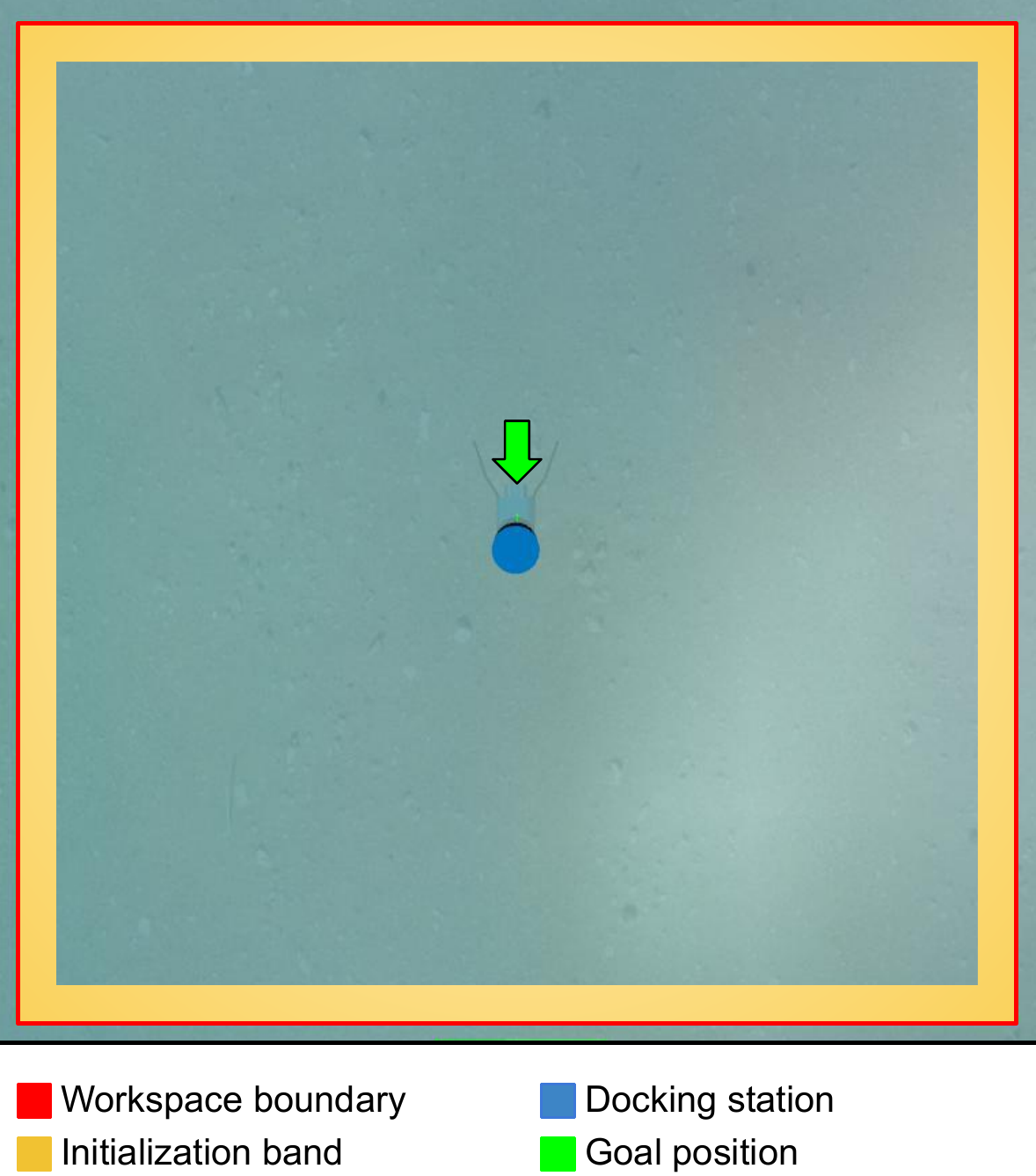}
    \caption{Figure illustrating the experimental setup.}
    \label{fig:experimentalsetup}
\end{figure}

In this work, we evaluate 3 state-of-the-art DRL algorithms, namely proximal policy optimization (PPO), twin delayed deep deterministic policy gradients (TD3) and soft actor-critic (SAC), using metrics such as average episodic rewards, average steps required to reach the docking station, success rate as well as demonstrations of the learned policies.
Our evaluations mitigate the bias of starting from a fixed initial state by using a random distribution of initial states that are scattered uniformly around the docking station within a 9m square workspace as shown in figure \ref{fig:experimentalsetup}.

The training was performed on a Xeon E5-2630 v3 CPU with 128GB ECC RAM and lasted roughly 16hrs, while using a 4 layer MLP network architecture with $[400, 300, 200, 100]$ neurons with ReLU activation for the hidden layers and tanh activation for the output layer(inspired by the work in \cite{henderson2017deep}).

We make use of the average episodic return as the performance metric of choice and supplement our results with a demonstration of the learned policy in action \cite{henderson2017deep}, to avoid any misleading conclusions.

\begin{table}[t]
\caption{Values used for all weights described in the reward function. These values were inspired by the work of \cite{Anderlini_2019}, \cite{rl-zoo} in the `Lunarlander' environment as well as our own evaluation.}
\label{tab:weightstable}
\centering
\begin{tabular}{lllllll} 
\toprule
	\textbf{}                                                                                          & \textbf{$w_d$} & \textbf{$w_{th_1}$} & \textbf{$w_{th_2}$} & \textbf{$w_{th_3}$} & \textbf{$w_\psi$} & \textbf{$w_y$} \\ 
\midrule
\begin{tabular}[c]{@{}l@{}}\textbf{Within the }\\\textbf{docking }\\\textbf{pyramid}\end{tabular}  & 30             & 2              & 5              & 5              & 1.3               & 1.2  \\ 
\midrule
\begin{tabular}[c]{@{}l@{}}\textbf{Outside the }\\\textbf{docking }\\\textbf{pyramid}\end{tabular} & 5              & 2              & 5              & 5              & 0                 & 0 \\
\bottomrule
\end{tabular}
\end{table}

\subsection{Reward Comparison}

\begin{figure}[t]
    \centering
    \includegraphics[scale=0.25]{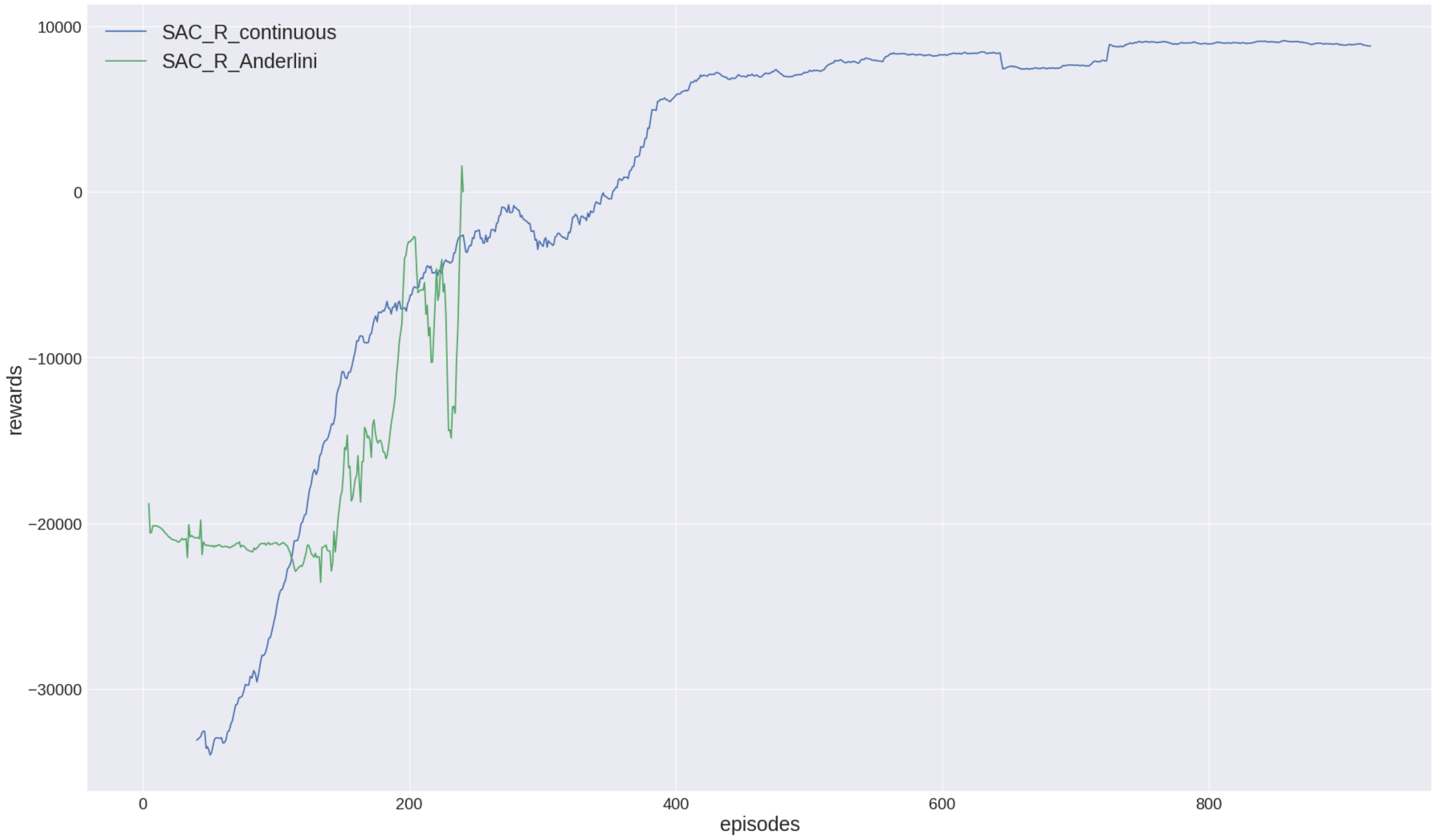}
    \caption{Comparison of average episodic returns of the SAC agents trained using the $\reward{anderlini}$ and $\reward{continuous}$ reward functions.}
    \label{fig:saccomp}
\end{figure}

First we perform a comparison between our reward function $\reward{continuous}$ shown in equation \ref{eqn:rewfunc} and the reward function described in \cite{Anderlini_2019} using the SAC algorithm as it's underlying entropy maximization formulation is the least likely to get stuck at a local optima. Also the authors of \cite{Anderlini_2019} suggest the usage of SAC to achieve better results in underwater docking. The SAC agents utilized in this evaluation conform to the default network architecture described in \ref{sec:setup}.

\begin{figure}[t]
    \centering
    \includegraphics[scale=0.3]{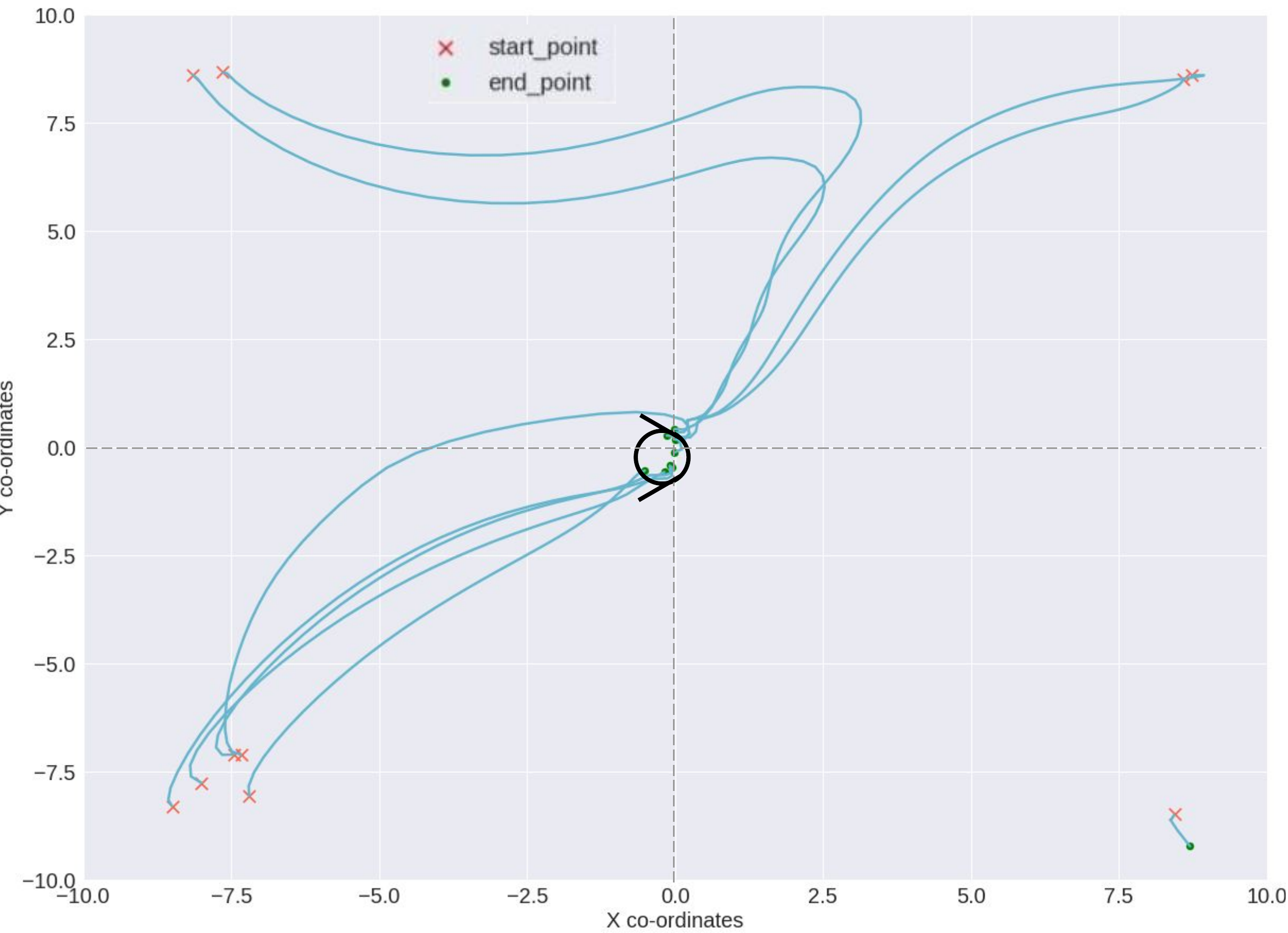}
    \caption{Behavior learned by the SAC agent when using the reward function $\reward{anderlini}$ (from \cite{Anderlini_2019}). The black circle is an approximate indication of the goal position along with it's tolerances.}
    \label{fig:sacpaper}
\end{figure}

Comparison of average episodic returns are shown in Figure \ref{fig:saccomp}. Our results indicate that our reward formulation works comparatively better than the one proposed by \cite{Anderlini_2019} this might be due to the fact that their reward function was primarily formulated for a 2D euclidean space encompassing the x-z axes, instead of the x-y axes.

In both cases the SAC agents were trained for $10^5$ total timesteps which translates to 932 episodes for $\reward{continuous}$ and 246 episodes for $\reward{anderlini}$ the large difference in number of episodes is due to the explicit penalization on the velocity component in $\reward{anderlini}$ which forces the AUV to move slowly towards the goal, hence taking up more steps per episode.

Our evaluation again consists of two runs of 5 episodes each, with episodes comprised of 150 timesteps within which the agent tries to reach the goal. Furthermore, comparing the behavior of two agents as shown in Figures \ref{fig:sacpaper} and \ref{fig:sac2d} we clearly see that the behavior learned via our reward function is vastly superior. The agents behavior observed in figure \ref{fig:sacpaper} not only indicates longer trajectories but it also shows that the agent was unable to reach the goal when the starting position of the AUV was behind the docking station in the fourth quadrant, which is in contrast to the behavior observed in \ref{fig:sac2d} where the agent is not only able to reach the goal from every quadrant but also does so with shorter trajectories.

Additionally the summary from Table \ref{tab:saccomp} shows that on average the reward function $\reward{anderlini}$ took longer to reach the goal pose, this is partly due to the explicit penalty placed on the AUV velocity in addition to penalizing the thruster utilization which considerably slows down the AUV movement. Since we do not explicitly penalize the velocity, instead doing it implicitly via the thruster penalization our reward function was able to drive the AUV towards the goal significantly faster.

\begin{table}[t]
\caption{Summary of evaluations performed on the reward functions $R_{Anderlini}$ and $R_{continuous}$. Here `Average steps to goal' represents the average number of timesteps required by the agent to reach the goal pose from a random starting point. Whereas `Succeeded in docking' refers to the number of times the agent was able to reach the goal point out of the possible 10 attempts.}
\label{tab:saccomp}    
\centering
\begin{tabular}{llll}
\toprule
\textbf{Reward Function} & \textbf{No. of}   & \textbf{Average steps} 	& \textbf{Succeeded} \\
                         & \textbf{Episodes} & \textbf{to goal}			& \textbf{in docking}\\
\midrule
$\reward{continuous}$ &  10 &  41.4 &  10 \\
$\reward{Anderlini}$ &  10 &  57.4 &  7 \\
\bottomrule
\end{tabular}
\end{table}

\subsection{Deep RL Algorithm Comparison}

\begin{figure}[t]
    \centering
    \includegraphics[scale=0.25]{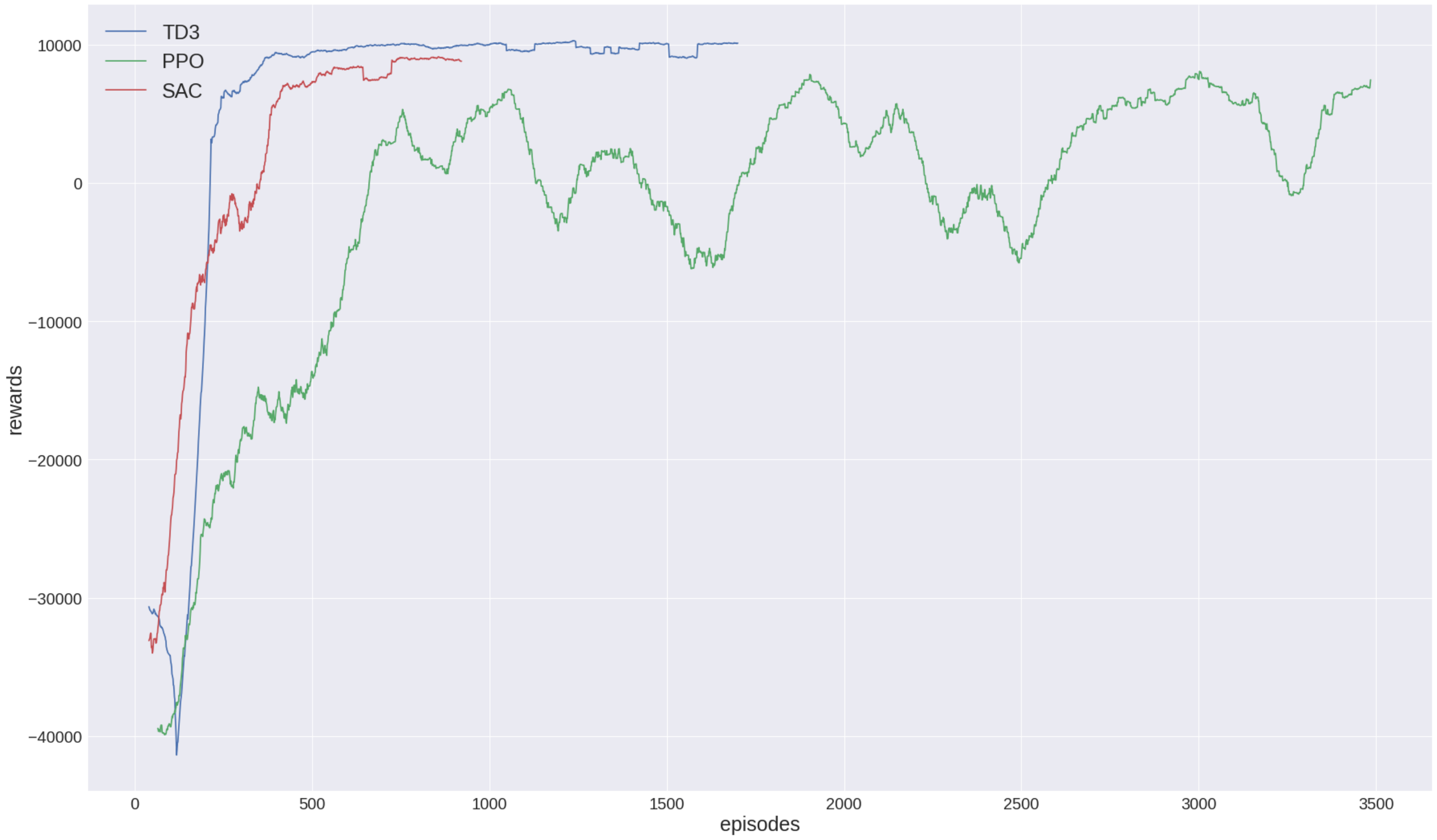}
    \caption{Comparison of average episodic returns of three agents trained respectively using the SAC, TD3 and PPO algorithms while utilizing the reward function $\reward{continuous}$ (from Eq. \ref{eqn:rewfunc}).}
    \label{fig:algocomp}
\end{figure}

In this section we perform the comparative evaluation between the three chosen Deep RL algorithms to find out which one performs the best for the 2D docking task along the x-y axes. We train all three agents for $10^5$ total timesteps and evaluate them over two test runs of 5 episodes each. We use our reward function $\reward{continuous}$ formulated in Eq \ref{eqn:rewfunc} along with the weights described in Table \ref{tab:weightstable}.
\label{sec:2d}

From figure \ref{fig:algocomp}, which presents a comparison of the average episodic returns for three agents (SAC, PPO, TD3), we conclude that the TD3 agent performs the best as it demonstrates the highest possible returns, with SAC coming in a close second while PPO performs the worst out of the three due to its issues with local optima.
Additionally, comparing the mean and standard deviation of the episodic rewards obtained by each agent during the two evaluation runs of five episodes as depicted in table \ref{tab:avgrew}, only reinforces the results seen in Figure \ref{fig:algocomp}. Although the mean of the episodic rewards for the SAC and TD3 agents are close, the standard deviation on the TD3 agent is considerably smaller than that of the SAC agent indicating that it is more consistent and reliable in finding its way to the goal position.

\begin{table}[t]
\caption{Average episodic returns and their standard deviations for the three chosen algorithms (PPO, SAC, and TD3).}
\label{tab:avgrew}    
\centering
\begin{tabular}{@{}llll@{}}
\toprule
\textbf{Algorithm}       & PPO                 & SAC                    & TD3\\
\midrule
\textbf{Average} & 328.2 $\pm 16949.3$ & 9413.613 $\pm 1574.2 $ & 10667.1 $\pm 688.8 $ \\
\textbf{Returns}\\
\bottomrule
\end{tabular}
\end{table}

\begin{figure}[t]
    \centering
    \includegraphics[scale=0.3]{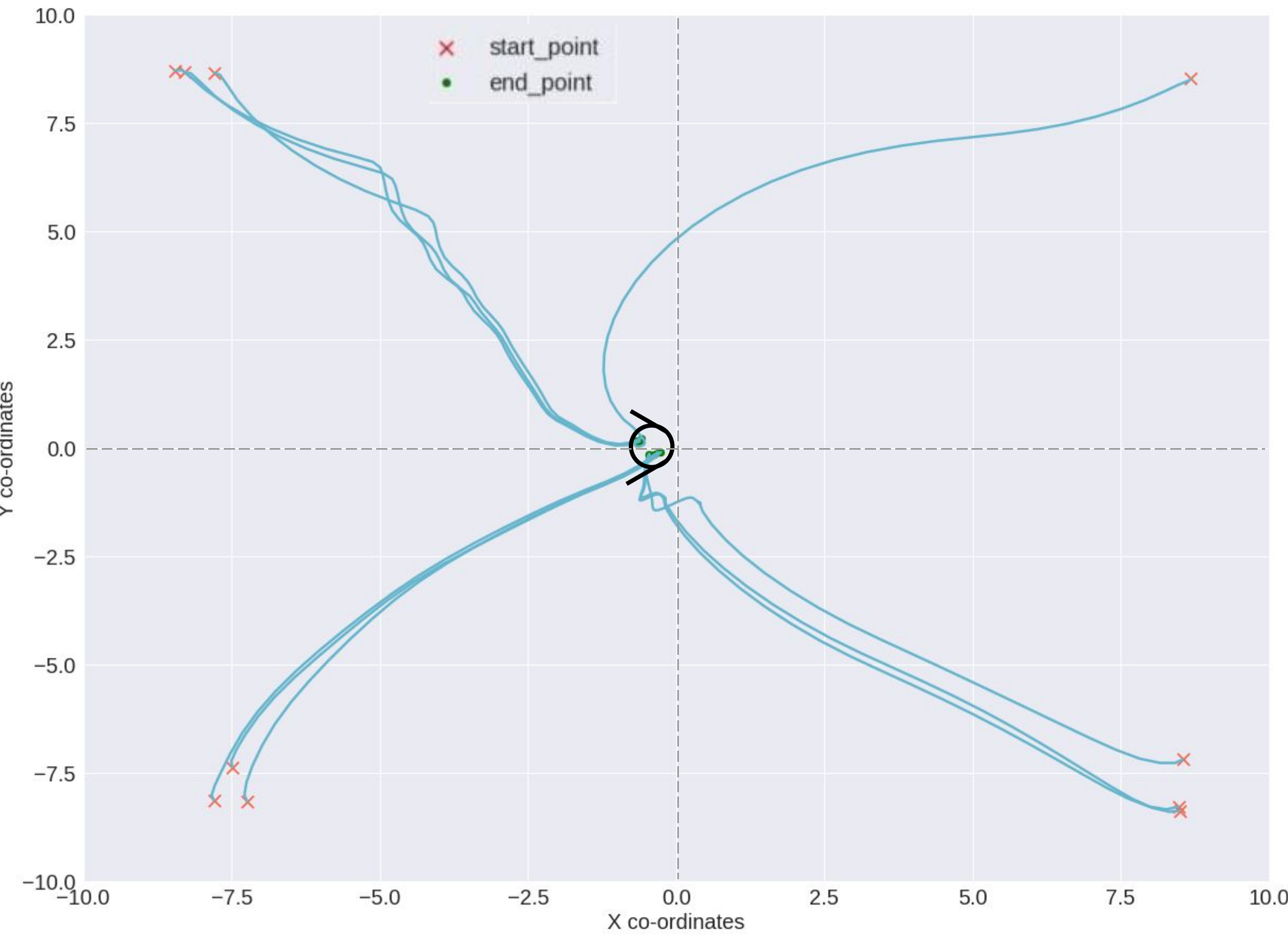}
    \caption{Behavior learned by the SAC agent when using the reward function $\reward{continuous}$ (from Eq \ref{eqn:rewfunc}). The black circle is an approximate indication of the goal position along with its tolerances.}
    \label{fig:sac2d}
\end{figure}

\begin{figure}[t]
    \centering
    \includegraphics[scale=0.3]{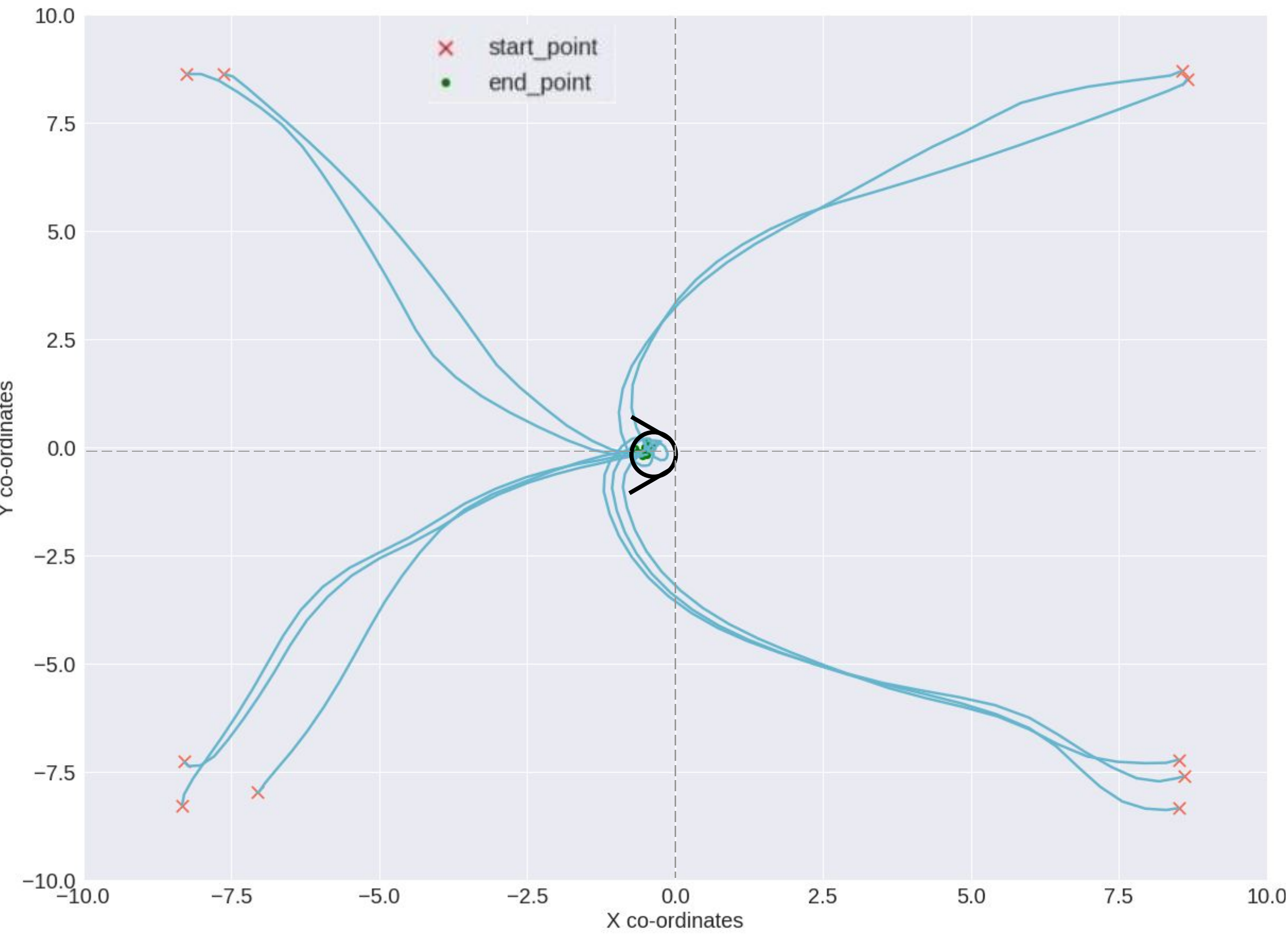}
    \caption{Behavior learned by the TD3 agent when using the reward function $\reward{continuous}$ (from Eq \ref{eqn:rewfunc}). The black circle is an approximate indication of the goal position along with its tolerances.}
    \label{fig:td32d}
\end{figure}

\begin{table}[t]
\caption{Summary of evaluation performed on three agents reward functions using the $\reward{continuous}$ reward function. Here `Average steps to goal' represents the average number of timesteps required by the agent to reach the goal pose from a random starting point. Whereas `Succeeded in docking' refers to the number of times the agent was able to reach the goal point out of the possible 10 attempts.}    
\centering
\begin{tabular}{llll}
\toprule
\textbf{Algorithm} 	& \textbf{No. of}   & \textbf{Average steps} 	& \textbf{Succeeded} \\
                    & \textbf{Episodes} & \textbf{to goal}			& \textbf{in docking}\\
\midrule
PPO & 10 & 35 & 6  \\
SAC & 10 & 41.4 & 10 \\
TD3 & 10 & 31   & 10 \\
\bottomrule
\end{tabular}
\label{tab:algocomp}
\end{table}

Moreover, upon comparing the behavior of the three agents as shown in Figures \ref{fig:sac2d} and \ref{fig:td32d} it becomes even clearer that the TD3 agent manages to outperform both the SAC and PPO agents. While both the SAC and TD3 agents were able to successfully solve the task of 2D docking when starting from any position in the four quadrants we observe that the trajectories of the TD3 agent are much smoother and smaller representing an efficient path to the goal pose. This is also reinforced from the data presented in Table \ref{tab:algocomp} where the TD3 agent was able to reach the goal with lesser number of timesteps per episode than the SAC agent. Thereby we conclude that the TD3 agent is the best suited for the task of 2D docking when utilizing the reward function $\reward{continuous}$.

\section{Conclusions and Future Work}
The work done in this paper demonstrates the performance of various model-free deep reinforcement learning algorithms for the task of underwater docking specifically in the two dimensional euclidean space. We also compare these algorithms on various reward functions that could be applied for the task of autonomous underwater docking. 

Our results show that the TD3 algorithm in combination with our reward formulation $\reward{continuous}$ achieves consistent and efficient docking of the DeepLeng AUV. Not only does it allow the agent to achieve higher episodic returns but is also able to do so without driving the AUV backwards towards the goal thus avoiding any unnecessary maneuvers, thus demonstrating smaller and quicker trajectories to the goal. The combination of our reward function with the SAC and TD3 algorithms successfully demonstrated this behavior even when starting from positions that lay behind the docking station. Thereby, conclusively making it the best suited reward function for the task of autonomous underwater docking that we evaluated.
    
We learned two lessons that we believe are important for the community. From evaluating the different reward functions mentioned in this work, orientation is a tricky component to control and should not be naively incorporated into the reward formulation. Cases where the orientation reward is present from the beginning as the agent starts its approach towards goal often exhibited sub-optimal maneuvers wherein the agent ended up driving the AUV backwards.

Another key takeaway would be to carefully formulate and test the reward function at every step of its development. Often it is the reward function that directs the agents behavior care should be put on what goes into it. An ill-formed reward function will not only hamper the learning process and result in sub-optimal behavior, but could also encourage the agent to exploit possible loopholes in order to attain the maximum possible rewards.

We hope the results achieved in this work can lead to better docking capabilities for AUVs and provide a concrete baseline to motivate further research into the field of autonomous underwater docking using techniques such as deep reinforcement learning or evolutionary algorithms. Our aim through this research was also to dispel some of the ambiguity that is associated with this topic especially due to the contentious nature of few studies that definitively explored this field.

Future work includes testing the learned policies in our AUV in a challenging environment, and to include additional sensor measurements for stability and robustness, including sonar sensing of the environment.

\section*{Acknowledgments}
This work was supported by the project EurEx-LUNa, funded by the German Ministry of Economic Affairs and Energy (BMWi) (Grant No. 50 NA 2002).

\bibliographystyle{ieeetr} 
\bibliography{bibliography.bib}
\end{document}